\pdfoutput=1

\documentclass[11pt]{article}

\usepackage[dvipsnames,table]{xcolor} 
\definecolor{DarkBlue}{RGB}{0,51,153} 
\usepackage[]{acl}

\usepackage{times}
\usepackage{latexsym}
\usepackage{bm}
\usepackage[T1]{fontenc}
\usepackage[LGR,T1]{fontenc} 
\usepackage[LGR,T1]{fontenc}
\usepackage[greek,english]{babel}
\usepackage[utf8]{inputenc}
\usepackage{adjustbox}
\usepackage{lscape}
\usepackage{graphicx}
\usepackage{multirow}

\usepackage{algorithm} 
\usepackage{algpseudocode} 
\usepackage{amssymb}

\usepackage{array} 
\usepackage{arydshln} 

\usepackage{microtype}

\title{FarFetched: Entity-centric Reasoning and Claim Validation for the Greek Language based on Textually Represented Environments}

\author{Dimitris Papadopoulos \\
  Technical University of Crete \\
  Chania, Greece \\
  \texttt{dpapadopoulos6@isc.tuc.gr} \\\ \And
  Katerina Metropoulou \\
  National Technical University of Athens\\
  Athens, Greece \\
  \texttt{kmetropoulou@mail.ntua.gr} \\\ \AND
  Nikolaos Matsatsinis \\
  Technical University of Crete \\
  Chania, Greece \\
  \texttt{nmatsatsinis@isc.tuc.gr} \\\ \And
  Nikolaos Papadakis \\
  Hellenic Army Academy \\
  Vari, Greece \\
  \texttt{npapadakis@sse.gr} \\}

\begin{document}
\maketitle
\begin{abstract}
Our collective attention span is shortened by the flood of online information. With \textit{FarFetched}, we address the need for automated claim validation based on the aggregated evidence derived from multiple online news sources. We introduce an entity-centric reasoning framework in which latent connections between events, actions, or statements are revealed via entity mentions and represented in a graph database. Using entity linking and semantic similarity, we offer a way for collecting and combining information from diverse sources in order to generate evidence relevant to the user's claim. Then, we leverage textual entailment recognition to quantitatively determine whether this assertion is credible, based on the created evidence. Our approach tries to fill the gap in automated claim validation for less-resourced languages and is showcased on the Greek language, complemented by the training of relevant semantic textual similarity (STS) and natural language inference (NLI) models that are evaluated on translated versions of common benchmarks. 
\end{abstract}

\section{Introduction}
\textbf{Motivation:} The wider diffusion of the Web since the dawn of Web 2.0 has enabled instantaneous access to an expanding universe of information. The entire nature of news consumption has shifted dramatically, as individuals increasingly rely on the Internet as their major source of information. While people access, filter and blend several websites into intricate patterns of media consumption, this wealth of information contained in billions of online articles inevitably creates a poverty of attention and a need to efficiently allocate this attention among the many sources that may absorb it. Verifying whether a given claim coheres with the knowledge hidden in the vast amount of published information is a fundamental problem in NLP, taking into account that the arrival of new information may weaken or retract the initially supported inference. The problem is more apparent in less-resourced languages that lack the necessary linguistic resources for building meaningful NLP applications. 

\begin{figure}[hbt!]
    \centering
        \includegraphics[scale=0.205]{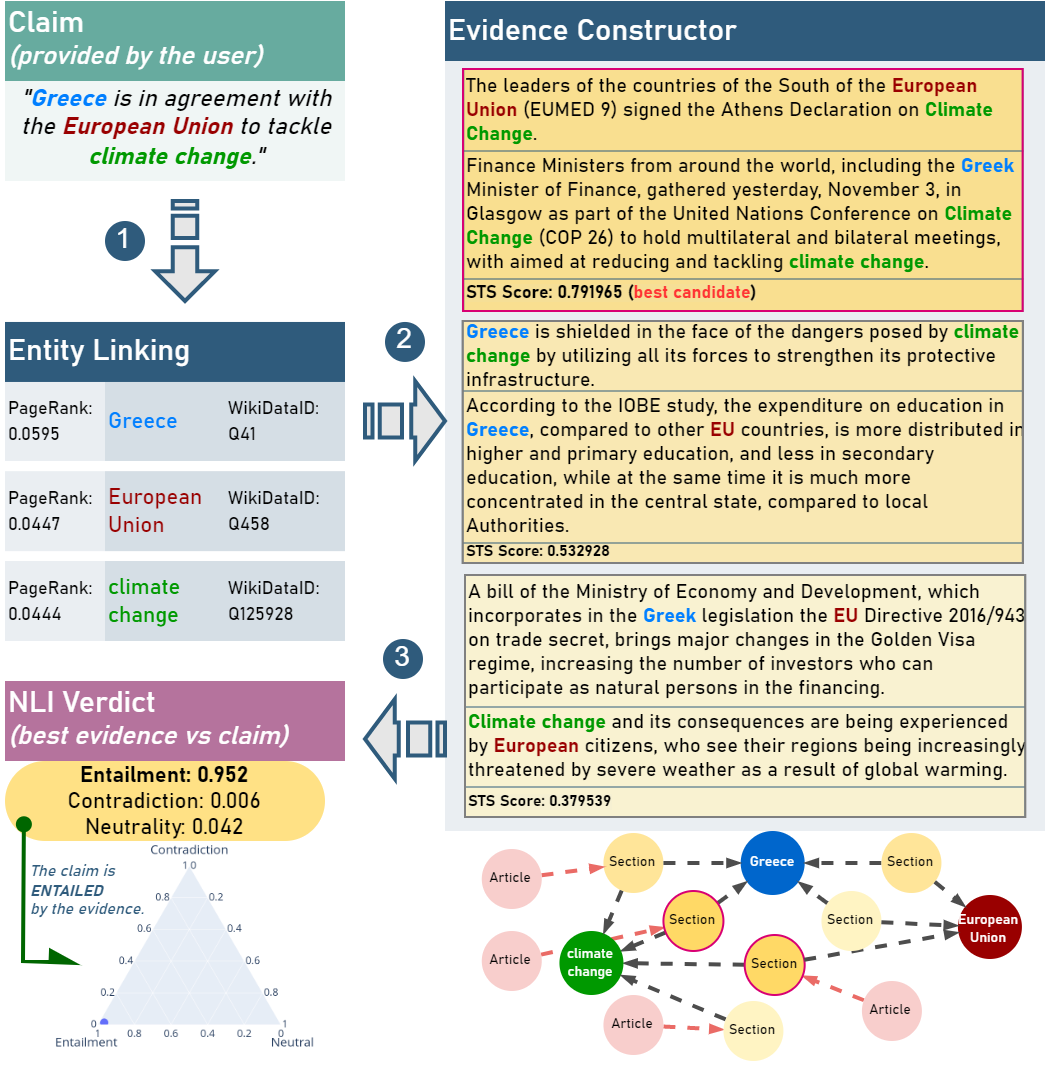}
        
    \caption{Claim validation example (translated from Greek) based on aggregated evidence using \textit{FarFetched}.}
    \label{fig:intro}
\end{figure}

\textbf{Approach and Contribution:} \textit{FarFetched} is a modular framework that enables people to verify any kind of textual claim based on the incorporated evidence from textual news sources. It combines a series of processes to periodically crawl for news articles and annotate their context with named entities. Given a user claim, \textit{FarFetched} derives a relevant subset of the stored content based on its semantic similarity with the provided claim, thus being able to reason about its validity in an NLI setting (Figure~\ref{fig:intro}). While the proposed framework focuses on the less-resourced Greek language, its modular architecture allows the integration of pretrained models for any language. Moreover, it is capable of topic-agnostic, evidence-aware assessment of arbitrary textual claims in a fully automated manner, without relying on feature engineering, curated sources and manual intervention.

The main contributions of this work are summarized as follows: a) to formalize, develop and evaluate a claim validation and reasoning approach based on the aggregated knowledge derived from the continuous monitoring of news sources, and b) to train, evaluate and share SotA models for the STS and NLI downstream tasks for the Greek language that support the core functionalities of our framework.\footnote{Code and benchmark datasets: \url{https://github.com/lighteternal/FarFetched_NLP}}

\section{Related Work}

Our work comprises functionalities comparable to those of fact checking frameworks, targeting the assignment of a truth value to a claim made in a particular context \cite{vlachos-riedel-2014-fact}. For most related approaches \cite{10.1145/3459637.3481985, majithia-etal-2019-claimportal, zhou-etal-2019-gear, ciampaglia2015computational, 10.1145/2463676.2463692} the evidence to support or refute a claim is derived from a trustworthy source (e.g. Wikipedia, crowdsourced tagging or expert annotators). Interesting deviations are DeClarE \cite{popat-etal-2018-declare} that searches for web articles related to a claim considering their in-between relevance using an attention mechanism, and ClaimEval \cite{Samadi_Talukdar_Veloso_Blum_2016}, based on first-order logic to contextualise prior knowledge from a set of the highest page-ranked websites. 

\textit{FarFetched} can be distinguished from the aforementioned works by four major points: a) evidence collection is disentangled from manual annotation but relies on a constantly updating feed of news articles instead; b) claim validation based on the accumulated evidence relies on the effective combination of entity linking and attention-based models; c) our approach provides interpretable reasoning based on the aggregated evidence of multiple sources without assessing their truthfulness as opposed to most fact checking frameworks; and d) the outcome of the process is dynamic as the continuous integration of new information may lead to a shift in the verdict of the validated claim. 

 Recent advances in the field of \textit{event-centric} NLP have introduced event representation methods based on narrative event chains \cite{vossen-etal-2015-storylines}, knowledge graphs \cite{tang-etal-2019-learning,vossen2016newsreader}, QA pairs \cite{michael-etal-2018-crowdsourcing} or event network embeddings \cite{zeng-etal-2021-gene} to capture connections among events in a global context. Our method relies on an \textit{entity-centric} approach instead, where the identified entities are used as connectors between events, actions, facts, statements or opinions, thus revealing latent connections between the articles containing them. A few similar approaches have been proposed for combining world knowledge with event extraction methods to represent coherent events, but rely either on causal reasoning to generate plausible predictions \cite{radinsky2012learning} or on QA models that require the accompanying news source to be provided along with the user's question \cite{jin-etal-2021-forecastqa}. 

The latest advances regarding the technological concepts that comprise our methodology are provided below:

Entity linking (EL) resolves the lexical ambiguity of entity mentions and determines their meanings in context. Typical EL approaches aim at identifying named entities in mention spans and linking them to entries of a KG (e.g. Wikidata, DBpedia) thus resolving their ambiguity. Recent methods combine the aforementioned tasks using local compatibility and topic similarity features \cite{delpeuch2019opentapioca}, pagerank-based wikification \cite{brank2017annotating} —used also in \textit{FarFetched}— or neural end-to-end models that jointly detect and disambiguate mentions with the help of context-aware mention embeddings \cite{kolitsas-etal-2018-end}.

The recent interest for encapsulating diverse semantic sentence features into fixed-size vectors has resulted in SotA systems for Semantic Textual Similarity (STS) based on supervised cross-sentence attention \cite{JMLR:v21:20-074}, Deep Averaging Networks (DAN) \cite{cer-etal-2018-universal} or siamese and triplet BERT-Networks \cite{reimers-gurevych-2019-sentence} to acquire meaningful sentence embeddings that can be compared using cosine similarity. The latter approach is leveraged in our case to train an STS model for the Greek language using transfer learning.

Finally, the task of Natural Language Inference (NLI) -also known as Recognizing Textual Entailment (RTE)- associates an input pair of premise and hypothesis phrases into one of three classes: contradiction, entailment and neutral. \citealp{ferreira-vlachos-2016-emergent} modeled fact checking as a form of RTE to predict whether a premise, typically part of a trusted source, is for, against, or observing a given claim. SotA NLI models typically rely on Transformer variants with global attention mechanisms \cite{DBLP:journals/corr/abs-2004-05150}, siamese network architectures \cite{reimers-gurevych-2019-sentence} (also used in \textit{FarFetched} to train a Greek NLI model), autoregressive language models for capturing long-term dependencies \cite{NEURIPS2019_dc6a7e65} and denoising autoencoders \cite{lewis-etal-2020-bart}.

\section{Methodology}

\subsection{Problem Definition}
Given a user claim in free text, we tackle the problem of deciding whether this statement is plausible based on the currently accumulated knowledge from news sources. We also acknowledge the problem of constructing relevant evidence from multiple sources by analysing the information contained in online articles and the need for efficiently extracting only contextually and semantically relevant excerpts to verify or refute the user's claim. While our work does not primarily focus on better sentence embeddings and natural language inference techniques, we also target the lack of such models for the Greek language. 

\subsection{Our approach}
\begin{figure}[hbt!]
    \centering
        \includegraphics[scale=0.25]{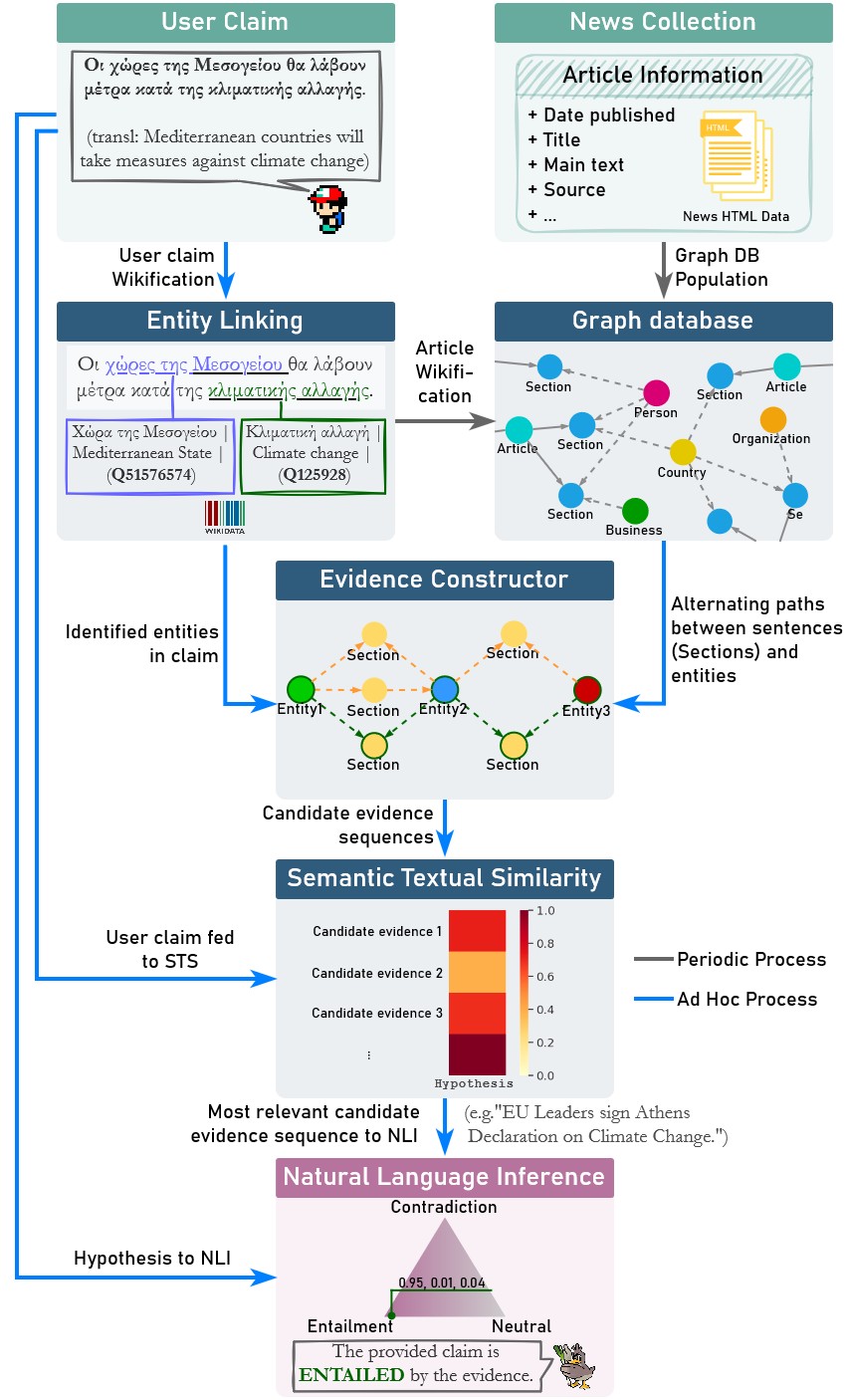}
        
    \caption{The \textit{FarFetched} modular framework.}
    \label{fig:farfetched}
\end{figure}

\textit{FarFetched} combines a series of \textit{offline} (i.e. performed periodically) operations to accumulate data from various news sources and annotate their context with named entities. It also encompasses a number of \textit{online} operations (i.e. upon user input) to assess the validity of a claim in free text. First, it identifies the entities included in the provided claim and leverages these as a starting point to derive a relevant subset of the stored textual information as candidate evidence. Each candidate is then compared with the claim in terms of textual similarity, in order to finally conclude on the most relevant evidence (premise) to reason about the validity of the claim (hypothesis) in an NLI setting. The distinct modules that comprise the framework are visualised in Figure~\ref{fig:farfetched}. The process that \textit{FarFetched} follows to evaluate a claim is summarized in Algorithm~\ref{alg:algo}, while each module is described in greater detail in the following subsections.

\begin{algorithm}[hbt!]\small
	\caption{\small Claim Evaluation}
	\label{alg:algo}
	 \hspace*{\algorithmicindent} \textbf{Input}: A claim $c$ provided by the user in natural language.  \\
     \hspace*{\algorithmicindent} \textbf{Output}: Most relevant evidence $seq^*$ (sequence of article excerpts)  based on the input claim $c$ along with its $STS\_score^*$ and $NLI\_score<c,e,n>$. \\
	\begin{algorithmic}[1]
		\State \textit{Entity Linking}: Find the set of entities ${(e_1,..., e_n) \in E}$, where $e\,\exists\, c$ and $\left| E \right| = n $
		\State $S \gets \varnothing$
		\State \textit{Graph database search}: Find all shortest paths $p$ between the alternating entities $e$ and sentences $s$: ${p \gets (e_1,s_a,e_2,...,e_{n-1},s_k,e_n) \in P}$
            \If{$P = \varnothing$}
                \State ${s \in P \iff s}$ has at least 1 entity mention
            \EndIf
		\For {${p_i \in P}$} 
		\State $seq_i \gets (s_a,...,s_k)$
		\State $S \gets S \cup seq_i$ (sequence $seq_i$ added to candidate evidence set)
		\EndFor
		\State $STS\_Scores \gets \varnothing$		
		\For {$seq_i \in S$} 
		\State \textit{Semantic Textual Similarity}: Compare $seq_i \in S$ to $c$ (each candidate evidence sequence to the claim) and calculate $STS\_score_i$	
		\State $STS\_Scores \gets STS\_Scores \cup STS\_score_i$		
		\EndFor
		\State Find the candidate $seq^*$ with the highest similarity to the claim: $STS\_score^* \gets max(STS\_Scores)$ \newline	 $seq^* \gets argmax(STS\_score^*)$	
		\State \textit{Natural Language Inference}: Compare  $seq^*$ to $c$ (the best candidate evidence to the claim) and calculate the scores for contradiction, entailment and neutrality $NLI\_score<c,e,n>$
	\end{algorithmic} 
\end{algorithm} 










\subsubsection{News Collection}
A multilingual, open-source crawler and extractor for heterogeneous website structures is leveraged to incorporate information from various news sources \cite{Hamborg2017}. It is capable of extracting the major properties of news articles (i.e., title, lead paragraph, main content, publication date, author, etc.), featuring full website extraction and requiring only the root URL of a news website to crawl it completely. 

\subsubsection{Graph Database Population}
The crawled articles are forwarded to a graph database \cite{10.1145/2384716.2384777} that initially stores only two types of nodes: \texttt{Article}, which represents a news article with its aforementioned properties and \texttt{Section} that represents a sentence of each article's main text (i.e. concatenated title and article body). Each \texttt{Article} node is linked to one or more \texttt{Section} nodes via the \texttt{HAS\_SECTION} relationship. 

\subsubsection{Entity Linking}
\begin{figure}[hbt!]
    \centering
        \includegraphics[scale=0.55]{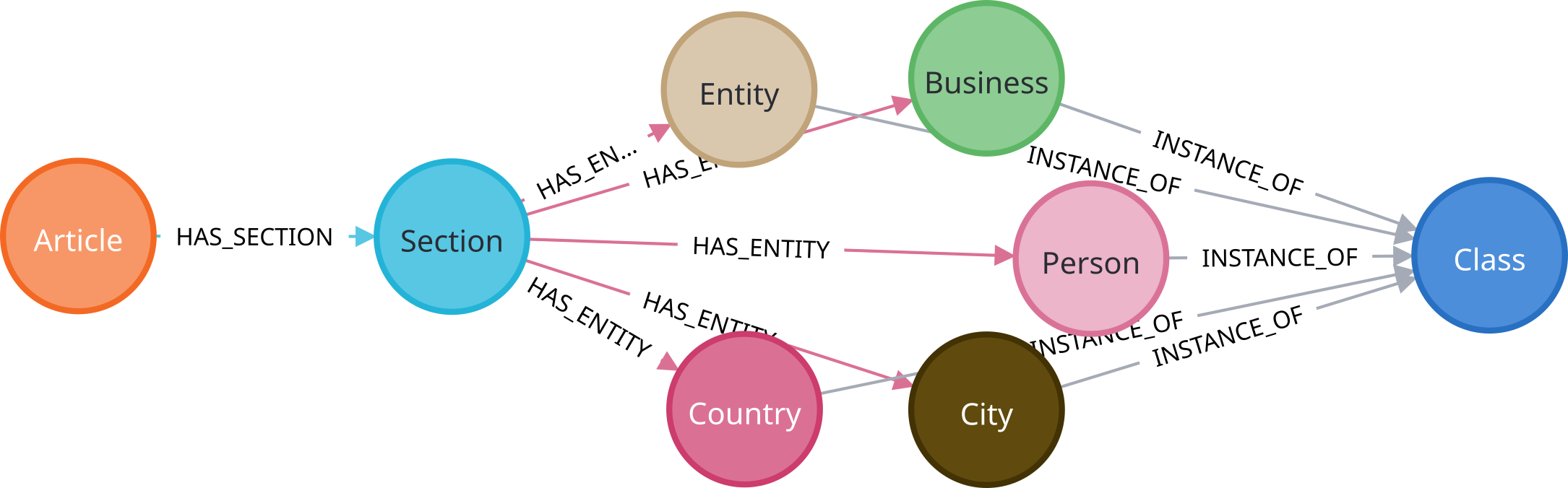}
        
    \caption{Final structure of the graph database.}
    \label{fig:metagraph.png}
\end{figure}

Given that our approach relies on largely unstructured textual documents that lack explicit semantic information, Entity Linking (EL) constitutes a central role in revealing latent connections between seemingly uncorrelated article sections. To this end, \textit{FarFetched} employs a type of semantic enrichment and entity disambiguation technique known as wikification \cite{JSIWikifier}, which involves using Wikipedia concepts as a source of semantic annotation. It applies pagerank-based wikification on input text to identify phrases that refer to entities of the target knowledge base (Wikipedia) and return their corresponding WikiData Entity ID. The latter is used as a unique identifier for storing the entities as \texttt{Entity} nodes to the graph database and for linking them with the crawled article \texttt{Section} nodes, resulting to a more tightly connected graph, where article sections are connected to WikiData entities via the \texttt{HAS\_ENTITY} relationship. 
The virtual graph of Figure~\ref{fig:metagraph.png} represents the structure (labels and relationships) of the graph database. 
It should be noted that an entity node might have an additional label (e.g. \texttt{Person}, \texttt{City}, \texttt{Business}) except for the generic \texttt{Entity} one, based on the WikiData class taxonomy.

\subsubsection{Evidence Constructor} \label{Evidence Constructor}
\begin{figure}[hbt!]
    \centering
        \includegraphics[scale=0.0390]{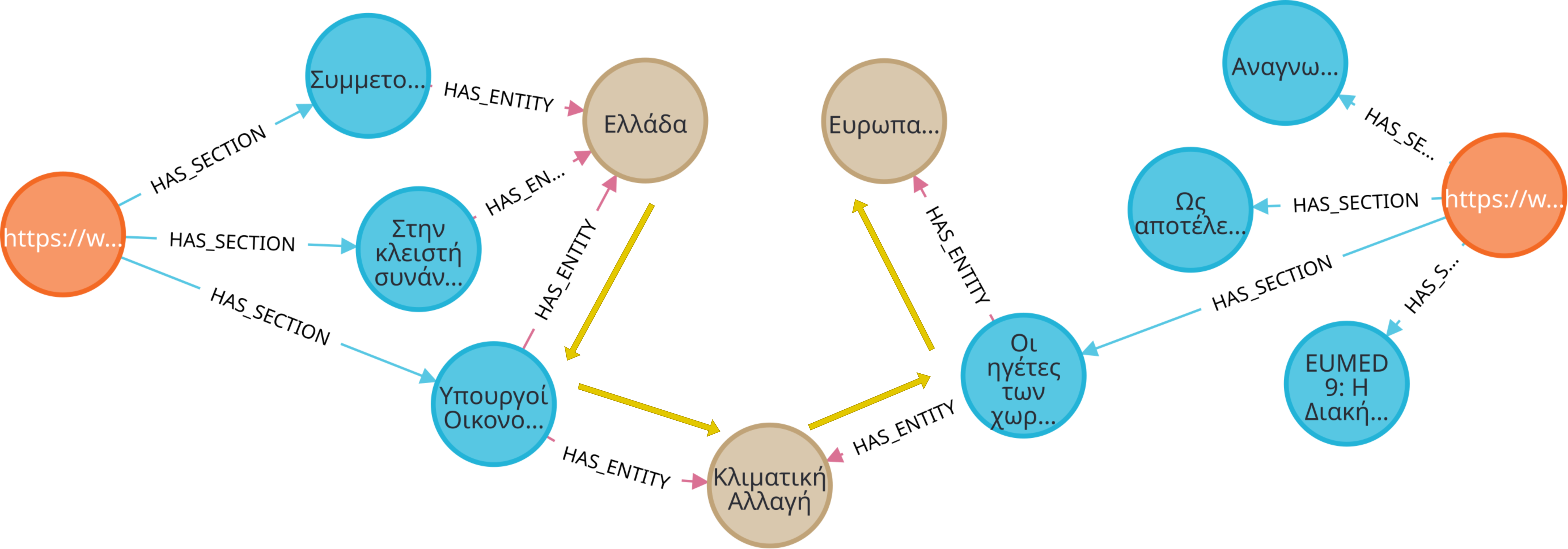}
        
    \caption{Shortest path example: The 3 entities (in brown) are connected with 2 article sections (in blue).}
    \label{fig:path-constructor}
\end{figure}

In a typical NLI setting, a premise represents our knowledge or evidence regarding an event and is used to infer whether a relevant hypothesis follows from it or not. In our case, article sections focusing on the same entities as the user's claim could potentially lead to the construction of useful evidence towards the validation of this claim. We can therefore leverage the entity-annotated article sections of our graph database to collect relevant evidence by aggregating information from multiple sources. To this end, we developed an evidence construction process that comprises the following steps:
\begin{enumerate}
\item The claim provided by the user passes through the Entity Linking phase and one or more entities (WikiData concepts) are identified.
\item The graph database is queried for all possible shortest paths that contain article sections between the identified entities. Given the implemented graph structure and $n$ Entity nodes, this translates to a minimum path length of $2(n-1)$ alternating Entity-Section nodes as shown in Figure~\ref{fig:path-constructor}. Since the existence of such path is not guaranteed, in cases that no path is found the algorithm will select an article section if it contains at least one mentioned entity.
\item The article sections contained in these paths are concatenated to form a set of candidate evidence sequences. Their relevance with the claim at hand is assessed during the Semantic Textual Similarity phase.
\end{enumerate}

\subsubsection{Semantic Textual Similarity}
We train and apply a sentence embeddings method to extract and compare the vector representations of the user's claim with each candidate evidence sequence, in order to select the most semantically relevant candidate for the final NLI phase. Despite the abundance of multilingual language models (e.g. m-BERT, XLM) that cover most common languages, pretrained multilingual sentence embeddings models do not generally perform well in downstream tasks for less-resourced languages like Greek \cite{10.1145/3411408.3411440}. Furthermore, given that the vector spaces between languages are not aligned, sentences with the same content in different languages could be mapped to different locations in the common vector space. To overcome this obstacle, we trained a Greek sentence embeddings model on parallel EN-EL (English-Greek) sentence pairs following a multilingual knowledge distillation approach \cite{reimers-gurevych-2020-making}. Our Greek student model ({XLM-RoBERTa}) was trained on parallel pairs to produce vectors for the EN-EL sentences that are close to the teacher's pretrained English model ones ({DistilRoBERTa}). Using the trained model, we are able to compare the produced vector representations between the claim and each concatenated candidate evidence sequence with regard to STS in terms of cosine similarity and forward the best candidate to the last phase of the claim validation process, namely Natural Language Inference.

\subsubsection{Natural Language Inference}
The last step of our process leverages NLI to determine whether the user claim (hypothesis) is entailed by, contradicted, or neutral to the most relevant evidence (premise) of the previous phase. To tackle the aforementioned multilinguality issues of pretrained language models on less-resourced languages, we finetuned a Greek \textit{sentence-transformers} Cross-Encoder \cite{reimers-gurevych-2019-sentence} (\textit{XLM-RoBERTa-base}) model for the NLI task. The model was trained on the Greek and English version of  the combined SNLI \cite{bowman-etal-2015-large} and MultiNLI \cite{williams-etal-2018-broad} corpora (AllNLI). We used the English-to-Greek machine translation model by \citealp{papadopoulos-etal-2021-penelopie} to create the Greek version of the AllNLI dataset. The trained model takes the premise-hypothesis pair as input and predicts one of the following labels for each case: "contradiction": c, "entailment": e or "neutral": n. The logits for each class are then converted to probabilities using the softmax function. These labels along with their probability scores can be used to assess whether the claim is verified by the accumulated knowledge of the candidate evidence.

\section{Experiments}

\subsection{Setup}

The technical details for each building block of \textit{FarFetched} are provided below:

\textbf{News Collection and Storage}:  The \textit{ news-please} \cite{Hamborg2017} Python library was used to ingest an initial corpus of news articles to support our experiments. The root URLs of two popular Greek news sites served as the starting point in order to recursively crawl news from a diverse topic spectrum, spanning from 2018 until 2021. We collected 13,236 articles, containing 31,358 sections in total. A\textit{ Neo4j} graph DBMS was used to store the crawled articles and sections as nodes and create their in-between relationships.

\textbf{Entity Linking}:  A Python script producing POST requests to the \textit{JSI Wikifier} web API \cite{brank2017annotating} was implemented to annotate the article sections and enrich the graph database with WikiData entities. A total of 2,516 WikiData entities of different types (e.g. sovereign states, cities, humans, organizations, academic institutions etc.) were identified in the crawled articles. A \textit{pageRankSqThreshold} of 0.80 was set for pruning the annotations on the basis of their pagerank score. 

\textbf{Evidence Constructor}:  We implemented Algorithm~\ref{alg:algo} as a Python script that executes a parametrizable Cypher query to construct candidate evidence sequences; the identified entities in the claim are used as parameters and the concatenated article sections that link these entities together are returned. For our experiments, the maximum number of relationships between the alternating Sections and Entities was set to $2(n-1)$ (shortest path), while the script returns candidate evidence sequences in descending order based on path length. These parameters can be modified if longer candidate evidence sequences are required.

\textbf{Semantic Similarity}: We finetuned a bilingual (Greek-English) \textit{XLM-RoBERTa-base} model (\textasciitilde270M parameters with 12-layers, 768-hidden-state, 3072 feed-forward hidden-state, 8-heads) using 340MB of parallel (EN-EL) sentences from various sources (e.g. OPUS, Wikimatrix, Tatoeba) leveraging the \textit{sentence-transformers} library \cite{reimers-2020-multilingual-sentence-bert}. The model was trained for 4 epochs with a batch size of 16 on a machine with a single NVIDIA GeForce RTX3080 (10GB of VRAM) for a total of 28 GPU-hours (single run).

\textbf{Natural Language Inference}: We finetuned a  Cross-Encoder \textit{XLM-RoBERTa-base} model of the same architecture on the created Greek-English AllNLI dataset (100MB) using \textit{sentence-transformers}. The model was trained on the same hardware setting for a single epoch, using a train batch size of 6 for 22 GPU-hours (single run).

\subsection{Main results}
In this section we perform a quantitative and qualitative demonstration of \textit{FarFetched}'s overall performance and also provide individual results for our STS and NLI models based on benchmark datasets.

\subsubsection{End-to-end performance}

Given the particularity of \textit{FarFetched} in evidence collection (data originating from constantly updating web content), a quantitative evaluation of its performance is quite challenging. To combat the lack of relevant benchmarks for the Greek language, we leveraged the FEVER dataset by \citealt{thorne-etal-2018-fever}, which models the assessment of truthfulness of written claims as a joint information retrieval and natural language inference task using evidence from Wikipedia. Each row of the dataset comprises a claim in free text, a list of evidence information including a URL to the Wikipedia page of the corresponding evidence and an annotated label (\texttt{SUPPORTS, REFUTES, NOT ENOUGH INFO}).
We manually translated a subset of 150 claims from the FEVER validation set from English to Greek and populated the graph database with the content of the corresponding Wikipedia URLs, which was automatically translated into Greek (due to its size), using the NMT model by \citealp{papadopoulos-etal-2021-penelopie}. We report \textit{FarFetched}'s performance in terms of accuracy, precision, recall and F1-score on Table \ref{tab:fever}.

\begin{table}[hbt!]
    \small
\centering
\label{tab:my-table}
\resizebox{\columnwidth}{!}{%
\begin{tabular}{c|ccc}
\hline
\textbf{Label}            & \multicolumn{1}{c|}{\textbf{Precision}} & \multicolumn{1}{c|}{\textbf{Recall}} & \textbf{F1-score} \\ \hline
\texttt{NOT ENOUGH INFO}                   & \multicolumn{1}{c|}{.36} & \multicolumn{1}{c|}{.80} & .49 \\
\texttt{REFUTES}                           & \multicolumn{1}{c|}{.91} & \multicolumn{1}{c|}{.72} & .80 \\
\texttt{SUPPORTS}                          & \multicolumn{1}{c|}{.84} & \multicolumn{1}{c|}{.70} & .76 \\
\textbf{Weighted Average} & \multicolumn{1}{c|}{\textbf{.82}}       & \multicolumn{1}{c|}{\textbf{.73}}    & \textbf{.75}      \\ \hline
\textbf{Label accuracy (overall)} & \multicolumn{3}{c}{\textbf{.73}}                          \\ \hline
\end{tabular}%
}
\caption{\textit{FarFetched} claim validation performance on Greek FEVER subset.}
\label{tab:fever}

\end{table}

The results indicate a balanced precision and recall for the \texttt{REFUTES} and \texttt{SUPPORTS} classes, while precision is relatively lower for the \texttt{NOT ENOUGH INFO} case. This can be partially attributed to the challenges of applying wikification on the automatically translated evidence content, leading to some claims not being linked to their corresponding evidence. 
Although the above results are not directly comparable to those of similar systems tested on the original English FEVER dataset, they show a significant gain over the baseline model of \citealt{thorne-etal-2018-fever} (label accuracy of 0.49). Based on a large comparative study conducted by \citealt{10.1145/3485127}, \textit{FarFetched} scores in the upper 30th percentile in terms of accuracy (scores ranging from 0.45 to 0.84); however, to the best of our knowledge none of these systems covers the Greek language.

We also provide a set of qualitative examples based on real data that aim at showcasing the capabilities of our system while also acknowledging the dynamicity of the evidence collection process. These scenarios are translated into English to facilitate readability. They include two parts each and are shown in Tables~\ref{tab:result1}, \ref{tab:result2} and \ref{tab:result3}. The original examples (in Greek) are available in the Appendix.

In \textit{Scenario 1}, two contradicting user claims (1a, 1b) with the same entity mentions are provided by the user (Table \ref{tab:result1}). Since they refer to the same entities, the Evidence Constructor returns the same candidate evidence sequences for both claims in order to evaluate their validity. The most relevant one (STS score in bold) is selected for the NLI phase, where the verdict is that the evidence entails the first claim (1a) and contradicts the second (1b).

\begin{table}[hbt!]
    \small
    \begin{tabular}{m{5.3cm}|>{\centering\arraybackslash}m{1.5cm}}
    
        \hline
        \textbf{User Claim \newline \scriptsize (Scenario 1)} & \textbf{NLI score} \\ \hline
        \textcolor{Mahogany}{Denmark} and \textcolor{DarkBlue}{Austria} believe that the \textcolor{teal}{European Union} should increase aid to refugees. (\textbf{1a})&
        \cellcolor{Dandelion!20} c: 0.014 \quad \textbf{e: 0.958} \quad n: 0.028 \\ \hline
        \textcolor{Mahogany}{Denmark} disagrees with \textcolor{DarkBlue}{Austria} on the management of immigration issues in the \textcolor{teal}{European Union}. (\textbf{1b}) &
        \cellcolor{Dandelion!20} \textbf{c: 0.951}\quad e: 0.002\quad n: 0.047 \\ \hline
            
        \multicolumn{2}{p{7.2cm}}{\textbf{Candidate Evidence Sequences ($\downarrow$ similarity)}} \\ \hline
        \multicolumn{2}{p{7.2cm}}{\cellcolor{Dandelion!20} \textcolor{DarkBlue}{Austria} and \textcolor{Mahogany}{Denmark} also want to increase \textcolor{teal}{EU} support for countries hosting refugees near crisis hotspots so that they do not travel to Europe. \newline {\scriptsize \textbullet \, \textbf{STS Score: 0.8505}}} \\ \hdashline
        \multicolumn{2}{p{7.2cm}}{\cellcolor{Dandelion!10} Checked by police at the Airport Police Departments ... the foreigners presented forged travel documents ... in order to leave the country for other \textcolor{teal}{EU} countries like France, Germany, Italy, \textcolor{DarkBlue}{Austria}, the Netherlands, \textcolor{Mahogany}{Denmark}, Spain and Norway. \newline {\scriptsize \textbullet \, STS Score: 0.2283}} \\ \hline
        
    \end{tabular}
    \caption{Demonstration of FarFetched on \textit{Scenario 1}.}
    \label{tab:result1}
\end{table}

\begin{figure}[hbt!]
    \centering
        \includegraphics[scale=0.13]{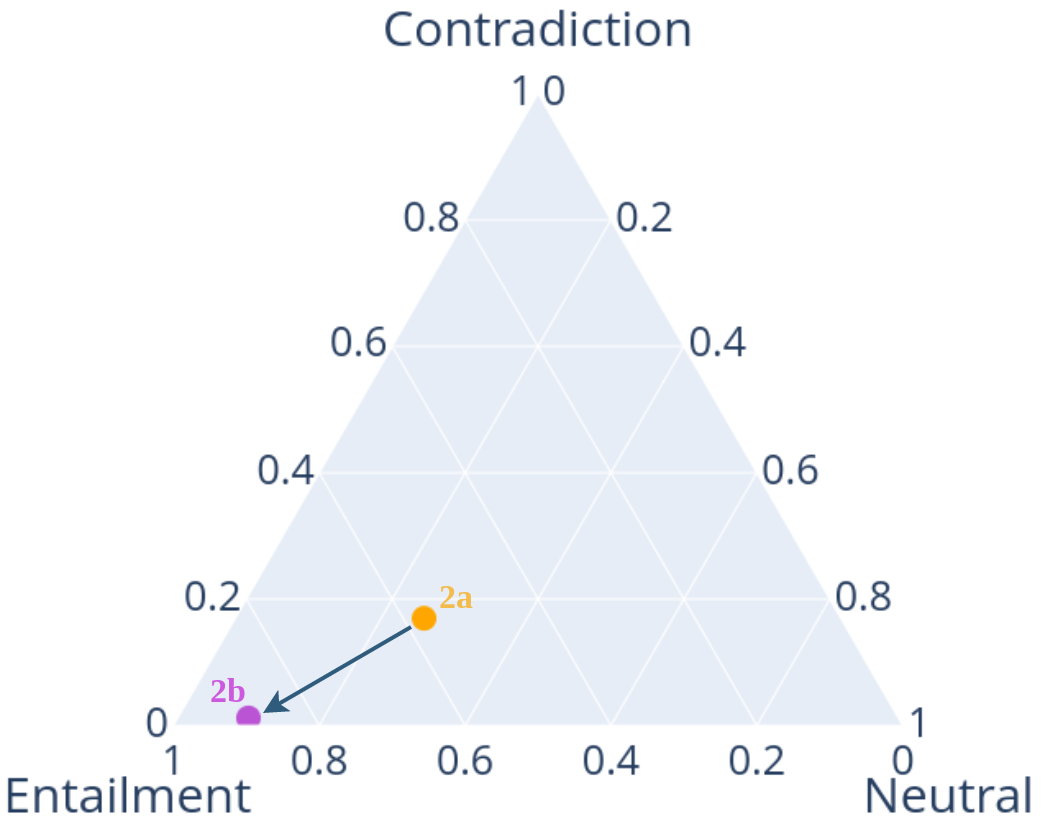}
        
    \caption{Shift in NLI verdict from Scenario 2a to Scenario 2b of Table \ref{tab:result2}.}
    \label{fig:nli_shift}
\end{figure}

In \textit{Scenario 2}, we investigate the sensitivity of our approach in exploiting new information to evaluate a claim (Table \ref{tab:result2}). The claim initially triggers the Evidence Constructor which returns multiple candidate evidence sequences, in descending STS order (yellow rows). During the NLI evaluation phase, the verdict is entailment, but with a low probability of 0.571 (2a). The same hypothesis is evaluated in Scenario 2b, after the addition of new information appended to the evidence list (blue row). The new evidence is clearly more relevant to the claim at hand, which is successfully identified by \textit{FarFetched}'s STS component that selects it as the best candidate, providing a more confident entailment score of 0.891 (2b). This shift in NLI verdict is visualized in Figure~\ref{fig:nli_shift}. Since \textit{FarFetched} relies on the constantly updating evidence, monitoring such shifts could be useful for identifying trend changes, especially for cases that benefit from long-term planning (business, market, politics etc.)

\begin{table}[hbt!]
    \small
    \begin{tabular}{m{3.4cm}|>{\centering\arraybackslash}m{1.5cm}|>{\centering\arraybackslash}m{1.5cm}}
        \hline
        
        \textbf{User Claim \newline \scriptsize (Scenario 2)} & \textbf{NLI score \scriptsize initial (2a)} & \textbf{NLI score \scriptsize updated (2b)} \\ \hline
        The \textcolor{Mahogany}{United States} plans to impose sanctions on \textcolor{DarkBlue}{Iran}. &
        \cellcolor{Dandelion!20} c: 0.170 \quad \textbf{e: 0.571} \quad n: 0.259 &
        \cellcolor{Orchid!20} c: 0.012 \quad \textbf{e: 0.891} \quad n: 0.097 \\ \hline
            
        \multicolumn{3}{p{7.2cm}}{\textbf{Candidate Evidence Sequences ($\downarrow$ similarity)}} \\ \hline
        \multicolumn{3}{p{7.2cm}}{\cellcolor{Dandelion!20}\textcolor{DarkBlue}{Iran} faces dilemma over whether to comply of Washington or will lead to collapse. The sanctions that came back in force today, will force the government of the Islamic Republic to accept the \textcolor{Mahogany}{US} claims regarding the Iranian nuclear program and Iranian activities in the Middle East East because, otherwise, the regime will be in danger to collapse, claimed Israel Kats, the Israeli minister responsible for Information Services. \newline {\scriptsize \textbullet \, \textbf{STS Score: 0.6665}}} \\ \hdashline
        \multicolumn{3}{p{7.2cm}}{\cellcolor{Dandelion!15} Why Greece was exempted from \textcolor{Mahogany}{US} sanctions on \textcolor{DarkBlue}{Iran}. New \textcolor{Mahogany}{US} sanctions on oil exports from \textcolor{DarkBlue}{Iran} have been  in force since November 5. \newline {\scriptsize \textbullet \, STS Score: 0.6324}} \\ \hdashline
        \multicolumn{3}{p{7.2cm}}{\cellcolor{Dandelion!10}"We are always in favor of diplomacy and talks ... But the Conversations need honesty ... The \textcolor{Mahogany}{US} is pushing again sanctions on \textcolor{DarkBlue}{Iran} and withdraw from the nuclear deal "(of 2015) and then they want to have conversations with us", Rohani said in a speech that was broadcast live on television. \newline {\scriptsize \textbullet \, STS Score: 0.5151}} \\ \hdashline
        \multicolumn{3}{p{7.2cm}}{\cellcolor{Orchid!20} \textbf{NEW:} Following the collapse of the last talks between the \textcolor{Mahogany}{US} and \textcolor{DarkBlue}{Iran}, the announcement of additional sanctions is expected in the coming days. \newline {\scriptsize \textbullet \, \textbf{STS Score: 0.7195}}} \\ \hline
        
    \end{tabular}
    \caption{Demonstration of FarFetched on \textit{Scenario 2}.}
    \label{tab:result2}
\end{table}

\textit{Scenario 3} is similar to \textit{2}, as one claim is evaluated on an initial set of candidate evidence sequences (3a) followed by a new relevant article section with contradicting evidence collected by the Evidence Collector in 3b (Table \ref{tab:result3}). However, in this case the new evidence is an excerpt from a person's interview. While our approach correctly identifies the relevance of this new evidence to the claim thus affecting the NLI verdict, it does not distinguish between opinions and factual evidence. This is discussed in more detail in Section \ref{Limitations}. 

\begin{table}[hbt!]
    \small
    \begin{tabular}{m{3.4cm}|>{\centering\arraybackslash}m{1.5cm}|>{\centering\arraybackslash}m{1.5cm}}
        \hline
        
        \textbf{User Claim \newline \scriptsize (Scenario 3)} & \textbf{NLI score \scriptsize initial (3a)} & \textbf{NLI score \scriptsize updated (3b)} \\ \hline
        \textcolor{Mahogany}{Apple} is trying to compete with \textcolor{DarkBlue}{Netflix} in the production of television content. &
        \cellcolor{Dandelion!20} c: 0.004 \quad \textbf{e: 0.967} \quad n: 0.029 &
        \cellcolor{Orchid!20} \textbf{c: 0.982} \quad e: 0.008 \quad n: 0.010 \\ \hline
        
        \multicolumn{3}{p{7.2cm}}{\textbf{Candidate Evidence Sequences ($\downarrow$ similarity)}} \\ \hline
        \multicolumn{3}{p{7.2cm}}{\cellcolor{Dandelion!20} \textcolor{Mahogany}{Apple} is expected to spend about \$ 2 billion this year creating original content that it hopes will compete with \textcolor{DarkBlue}{Netflix}, Hulu and Amazon, already established in the television audience. \newline {\scriptsize \textbullet \, \textbf{STS Score: 0.7107}}} \\ \hdashline
        \multicolumn{3}{p{7.2cm}}{ \cellcolor{Orchid!20} \textbf{NEW: }"We're not trying to compete with \textcolor{DarkBlue}{Netflix} on TV," an \textcolor{Mahogany}{Apple} spokesman said in an interview. \newline {\scriptsize \textbullet \, \textbf{STS Score: 0.7134}}} \\ \hline
        
    \end{tabular}
    \caption{Demonstration of FarFetched on \textit{Scenario 3}.}
    \label{tab:result3}
\end{table}

\subsubsection{STS performance}
The performance of our semantic similarity model was evaluated on the test subset of the STS2017 benchmark dataset \cite{cer-etal-2017-semeval}. Given that the original dataset does not provide sentence pairs in Greek, we manually created a cross-lingual version for the English-Greek pair. The performance is measured using Pearson (r) and Spearman ($\rho$) correlation between the predicted and gold similarity scores (Table~\ref{sts-table}). We also provide results regarding translation matching accuracy, evaluating the source and target language embeddings in terms of cosine similarity. Our model achieves a slightly better performance in both evaluations compared to the current state-of-the-art multilingual model by \citealp{reimers-gurevych-2019-sentence}. 

\begin{table}[hbt!]
    \small
    \begin{tabular}{m{2.85cm}|m{0.45cm}m{0.45cm}|m{0.8cm}m{0.8cm}}
        \hline
        
        \multirow{2}{2.85cm}{\vfil \textbf{Model}} & \multicolumn{2}{m{1.4cm}|}{STS2017} & \multicolumn{2}{m{2.1cm}}{Translation Matching} \\ \cline{2-5} 
        \multicolumn{1}{c|}{} & \textbf{r} & \boldsymbol{$\rho$} & \textbf{Acc. (en2el)} & \textbf{Acc. (el2en)} \\ \hline
        
        \textit{\textbf{STS-XLM-RoBERTa-base (Ours)}} & \textbf{83.30} & \textbf{84.32} & \textbf{98.05} & \textbf{97.80} \\
        Paraphrase-multilingual-mpnet-base-v2 (UKP-TUDA) & 82.71 & \multicolumn{1}{c|}{82.70} & 97.50 & 97.35 \\ \hline
        
    \end{tabular}
    \caption{\label{sts-table}
    STS model comparison on EN-EL version of STS2017 and in terms of translation matching accuracy.}
\end{table}

\subsubsection{NLI performance}
We benchmark our trained NLI model on the Greek subset of the XNLI dataset \cite{conneau-etal-2018-xnli} that contains 5,010 premise-hypothesis pairs (Table~\ref{nli-table}). Despite not having used the XNLI dataset during the training phase, we achieve a 1\% gain over the multilingual XLM-R \cite{conneau-etal-2020-unsupervised} and are on par with the monolingual Greek-BERT by \citealp{10.1145/3411408.3411440}. Since our model was trained on a mixture of Greek and English sentence pairs, it is more suitable for corpora that also contain English terms (e.g. technology, science topics) without suffering from the under-representability of the Greek language occurring in multilingual models.

\begin{table}[hbt!]
    \small
\centering
\begin{tabular}{l|l}
\hline
\textbf{Model}                           & \textbf{F1-score} \\ \hline
\textit{\textbf{NLI-XLM-RoBERTa-base (Ours)}} & \textbf{78.3}     \\
Greek-BERT (AUEB)                        & 78.6 ± 0.62       \\
XLM-RoBERTa-base (Facebook)          & 77.3 ± 0.41       \\
M-BERT (Google AI Language)           & 73.5 ± 0.49       \\ \hline
\end{tabular}
\caption{\label{nli-table}
NLI model comparison in terms of F1-score on the Greek subset of XNLI-test dataset.}
\end{table}

\section{Limitations} \label{Limitations}
We acknowledge that \textit{FarFetched} is possible to encounter errors in 3 main areas; these limitations are briefly addressed below.

\textbf{Entity Linking}: Highly ambiguous entities and name variations pose challenges to any entity linking method. Since we claim that our approach is entity-centric, a wrong entity annotation may lead to irrelevant candidate evidence sequences and increase the probability of "neutral" NLI verdicts. Moreover, the tunable sensitivity of the integrated wikification module implies a trade-off between a precision-oriented and a recall-oriented strategy, the latter resulting in more annotated articles, but also being prone to false-positive annotations.  

\textbf{Evidence Construction}: This initial version of our approach relies solely on the STS comparison between the evidence and the claim, based on a shortest path approach as discussed in Section \ref{Evidence Constructor}. In cases that involve a larger number of entities in the user claim, calculating the shortest path between the alternating Entity-Section nodes can be computationally cumbersome. Moreover, there is no guarantee that the shortest path is able to capture the most relevant candidate evidence sequences; to this end, outputting the top $n$ best candidates is considered, providing a user with an overview of the extracted news excerpts together with their NLI outcome. Finally, neither a temporal evaluation of the evidence with regard to the claim nor a distinction between opinions and facts is considered; all candidates are treated as equal. 

\textbf{Natural Language Inference}: Recognizing the entailment between a pair of sentences partially depends on the tense and aspect of the predications. Tense plays an important role in determining the temporal location of the predication (i.e. past, present or future), while the aspectual auxiliaries signify an event's internal constituency (e.g. whether an action is completed or in progress). While the work of \citealp{kober-etal-2019-temporal} indicates that language models substantially encode morphosyntactic information regarding tense and aspect, they are unable to reason based only on these properties. To this end, claims with a high presence of such semantic properties should be avoided.  

\section{Conclusions}
In this work, we presented a novel approach for claim validation and reasoning based on the accumulated knowledge from the continuous ingestion and processing of news articles. \textit{FarFetched} is able to evaluate the validity of any arbitrary textual claim by automatically retrieving and aggregating evidence from multiple sources, relying on the pillars of entity linking, semantic textual similarity and natural language inference.

We showcased the effectiveness of our method on the FEVER benchmark as well as on diverse scenarios and acknowledged its limitations. As byproducts of our work, we trained and open-sourced an NLI and an STS model for the less-resourced Greek language, achieving state-of-the-art performance on the XNLI and STS2017 benchmarks respectively. While our framework fills the gap in automated claim validation for Greek, its modular architecture allows it to be repurposed for any language for which the corresponding models exist.

For future work, we intend to address the limitations of our method mentioned in Section \ref{Limitations}, focusing primarily on an optimal entity linking setting, as well as on a more robust strategy for constructing relevant candidate evidence sequences.

\section*{Acknowledgments}

The research work of Dimitris Papadopoulos was supported by the Hellenic Foundation for Research and Innovation (HFRI) under the HFRI PhD Fellowship grant (Fellowship Number: 50, 2nd call).

\bibliography{anthology,custom}

\begin{thebibliography}{40}
\expandafter\ifx\csname natexlab\endcsname\relax\def\natexlab#1{#1}\fi

\bibitem[{Bekoulis et~al.(2021)Bekoulis, Papagiannopoulou, and
  Deligiannis}]{10.1145/3485127}
Giannis Bekoulis, Christina Papagiannopoulou, and Nikos Deligiannis. 2021.
\newblock \href {https://doi.org/10.1145/3485127} {A review on fact extraction
  and verification}.
\newblock \emph{ACM Comput. Surv.}, 55(1).

\bibitem[{Beltagy et~al.(2020)Beltagy, Peters, and
  Cohan}]{DBLP:journals/corr/abs-2004-05150}
Iz~Beltagy, Matthew~E. Peters, and Arman Cohan. 2020.
\newblock \href {http://arxiv.org/abs/2004.05150} {Longformer: The
  long-document transformer}.
\newblock \emph{CoRR}, abs/2004.05150.

\bibitem[{Bowman et~al.(2015)Bowman, Angeli, Potts, and
  Manning}]{bowman-etal-2015-large}
Samuel~R. Bowman, Gabor Angeli, Christopher Potts, and Christopher~D. Manning.
  2015.
\newblock \href {https://doi.org/10.18653/v1/D15-1075} {A large annotated
  corpus for learning natural language inference}.
\newblock In \emph{Proceedings of the 2015 Conference on Empirical Methods in
  Natural Language Processing}, pages 632--642, Lisbon, Portugal. Association
  for Computational Linguistics.

\bibitem[{Brank et~al.(2017{\natexlab{a}})Brank, Leban, and
  Grobelnik}]{brank2017annotating}
Janez Brank, Gregor Leban, and Marko Grobelnik. 2017{\natexlab{a}}.
\newblock Annotating documents with relevant wikipedia concepts.
\newblock \emph{Proceedings of SiKDD}.

\bibitem[{Brank et~al.(2017{\natexlab{b}})Brank, Leban, and
  Grobelnik}]{JSIWikifier}
Janez Brank, Gregor Leban, and Marko Grobelnik. 2017{\natexlab{b}}.
\newblock Annotating documents with relevant wikipedia concepts.
\newblock In \emph{Proceedings of the Slovenian Conference on Data Mining and
  Data Warehouses (SiKDD 2017)}, pages 218--223.

\bibitem[{Cer et~al.(2017)Cer, Diab, Agirre, Lopez-Gazpio, and
  Specia}]{cer-etal-2017-semeval}
Daniel Cer, Mona Diab, Eneko Agirre, I{\~n}igo Lopez-Gazpio, and Lucia Specia.
  2017.
\newblock \href {https://doi.org/10.18653/v1/S17-2001} {{S}em{E}val-2017 task
  1: Semantic textual similarity multilingual and crosslingual focused
  evaluation}.
\newblock In \emph{Proceedings of the 11th International Workshop on Semantic
  Evaluation ({S}em{E}val-2017)}, pages 1--14, Vancouver, Canada. Association
  for Computational Linguistics.

\bibitem[{Cer et~al.(2018)Cer, Yang, Kong, Hua, Limtiaco, St.~John, Constant,
  Guajardo-Cespedes, Yuan, Tar, Strope, and Kurzweil}]{cer-etal-2018-universal}
Daniel Cer, Yinfei Yang, Sheng-yi Kong, Nan Hua, Nicole Limtiaco, Rhomni
  St.~John, Noah Constant, Mario Guajardo-Cespedes, Steve Yuan, Chris Tar,
  Brian Strope, and Ray Kurzweil. 2018.
\newblock \href {https://doi.org/10.18653/v1/D18-2029} {Universal sentence
  encoder for {E}nglish}.
\newblock In \emph{Proceedings of the 2018 Conference on Empirical Methods in
  Natural Language Processing: System Demonstrations}, pages 169--174,
  Brussels, Belgium. Association for Computational Linguistics.

\bibitem[{Ciampaglia et~al.(2015)Ciampaglia, Shiralkar, Rocha, Bollen, Menczer,
  and Flammini}]{ciampaglia2015computational}
Giovanni~Luca Ciampaglia, Prashant Shiralkar, Luis~M Rocha, Johan Bollen,
  Filippo Menczer, and Alessandro Flammini. 2015.
\newblock Computational fact checking from knowledge networks.
\newblock \emph{PloS one}, 10(6):e0128193.

\bibitem[{Conneau et~al.(2020)Conneau, Khandelwal, Goyal, Chaudhary, Wenzek,
  Guzm{\'a}n, Grave, Ott, Zettlemoyer, and
  Stoyanov}]{conneau-etal-2020-unsupervised}
Alexis Conneau, Kartikay Khandelwal, Naman Goyal, Vishrav Chaudhary, Guillaume
  Wenzek, Francisco Guzm{\'a}n, Edouard Grave, Myle Ott, Luke Zettlemoyer, and
  Veselin Stoyanov. 2020.
\newblock \href {https://doi.org/10.18653/v1/2020.acl-main.747} {Unsupervised
  cross-lingual representation learning at scale}.
\newblock In \emph{Proceedings of the 58th Annual Meeting of the Association
  for Computational Linguistics}, pages 8440--8451, Online. Association for
  Computational Linguistics.

\bibitem[{Conneau et~al.(2018)Conneau, Rinott, Lample, Williams, Bowman,
  Schwenk, and Stoyanov}]{conneau-etal-2018-xnli}
Alexis Conneau, Ruty Rinott, Guillaume Lample, Adina Williams, Samuel Bowman,
  Holger Schwenk, and Veselin Stoyanov. 2018.
\newblock \href {https://doi.org/10.18653/v1/D18-1269} {{XNLI}: Evaluating
  cross-lingual sentence representations}.
\newblock In \emph{Proceedings of the 2018 Conference on Empirical Methods in
  Natural Language Processing}, pages 2475--2485, Brussels, Belgium.
  Association for Computational Linguistics.

\bibitem[{Delpeuch(2019)}]{delpeuch2019opentapioca}
Antonin Delpeuch. 2019.
\newblock Opentapioca: Lightweight entity linking for wikidata.
\newblock \emph{arXiv preprint arXiv:1904.09131}.

\bibitem[{Ferreira and Vlachos(2016)}]{ferreira-vlachos-2016-emergent}
William Ferreira and Andreas Vlachos. 2016.
\newblock \href {https://doi.org/10.18653/v1/N16-1138} {{E}mergent: a novel
  data-set for stance classification}.
\newblock In \emph{Proceedings of the 2016 Conference of the North {A}merican
  Chapter of the Association for Computational Linguistics: Human Language
  Technologies}, pages 1163--1168, San Diego, California. Association for
  Computational Linguistics.

\bibitem[{Goasdou\'{e} et~al.(2013)Goasdou\'{e}, Karanasos, Katsis, Leblay,
  Manolescu, and Zampetakis}]{10.1145/2463676.2463692}
Fran\c{c}ois Goasdou\'{e}, Konstantinos Karanasos, Yannis Katsis, Julien
  Leblay, Ioana Manolescu, and Stamatis Zampetakis. 2013.
\newblock \href {https://doi.org/10.1145/2463676.2463692} {Fact checking and
  analyzing the web}.
\newblock In \emph{Proceedings of the 2013 ACM SIGMOD International Conference
  on Management of Data}, SIGMOD '13, page 997–1000, New York, NY, USA.
  Association for Computing Machinery.

\bibitem[{Hamborg et~al.(2017)Hamborg, Meuschke, Breitinger, and
  Gipp}]{Hamborg2017}
Felix Hamborg, Norman Meuschke, Corinna Breitinger, and Bela Gipp. 2017.
\newblock \href {https://doi.org/10.5281/zenodo.4120316} {news-please: A
  generic news crawler and extractor}.
\newblock In \emph{Proceedings of the 15th International Symposium of
  Information Science}, pages 218--223.

\bibitem[{Jin et~al.(2021)Jin, Khanna, Kim, Lee, Morstatter, Galstyan, and
  Ren}]{jin-etal-2021-forecastqa}
Woojeong Jin, Rahul Khanna, Suji Kim, Dong-Ho Lee, Fred Morstatter, Aram
  Galstyan, and Xiang Ren. 2021.
\newblock \href {https://doi.org/10.18653/v1/2021.acl-long.357}
  {{F}orecast{QA}: A question answering challenge for event forecasting with
  temporal text data}.
\newblock In \emph{Proceedings of the 59th Annual Meeting of the Association
  for Computational Linguistics and the 11th International Joint Conference on
  Natural Language Processing (Volume 1: Long Papers)}, pages 4636--4650,
  Online. Association for Computational Linguistics.

\bibitem[{Kober et~al.(2019)Kober, Bijl~de Vroe, and
  Steedman}]{kober-etal-2019-temporal}
Thomas Kober, Sander Bijl~de Vroe, and Mark Steedman. 2019.
\newblock \href {https://doi.org/10.18653/v1/W19-0409} {Temporal and aspectual
  entailment}.
\newblock In \emph{Proceedings of the 13th International Conference on
  Computational Semantics - Long Papers}, pages 103--119, Gothenburg, Sweden.
  Association for Computational Linguistics.

\bibitem[{Kolitsas et~al.(2018)Kolitsas, Ganea, and
  Hofmann}]{kolitsas-etal-2018-end}
Nikolaos Kolitsas, Octavian-Eugen Ganea, and Thomas Hofmann. 2018.
\newblock \href {https://doi.org/10.18653/v1/K18-1050} {End-to-end neural
  entity linking}.
\newblock In \emph{Proceedings of the 22nd Conference on Computational Natural
  Language Learning}, pages 519--529, Brussels, Belgium. Association for
  Computational Linguistics.

\bibitem[{Koutsikakis et~al.(2020)Koutsikakis, Chalkidis, Malakasiotis, and
  Androutsopoulos}]{10.1145/3411408.3411440}
John Koutsikakis, Ilias Chalkidis, Prodromos Malakasiotis, and Ion
  Androutsopoulos. 2020.
\newblock \href {https://doi.org/10.1145/3411408.3411440} {Greek-bert: The
  greeks visiting sesame street}.
\newblock In \emph{11th Hellenic Conference on Artificial Intelligence}, SETN
  2020, page 110–117, New York, NY, USA. Association for Computing Machinery.

\bibitem[{Lewis et~al.(2020)Lewis, Liu, Goyal, Ghazvininejad, Mohamed, Levy,
  Stoyanov, and Zettlemoyer}]{lewis-etal-2020-bart}
Mike Lewis, Yinhan Liu, Naman Goyal, Marjan Ghazvininejad, Abdelrahman Mohamed,
  Omer Levy, Veselin Stoyanov, and Luke Zettlemoyer. 2020.
\newblock \href {https://doi.org/10.18653/v1/2020.acl-main.703} {{BART}:
  Denoising sequence-to-sequence pre-training for natural language generation,
  translation, and comprehension}.
\newblock In \emph{Proceedings of the 58th Annual Meeting of the Association
  for Computational Linguistics}, pages 7871--7880, Online. Association for
  Computational Linguistics.

\bibitem[{Majithia et~al.(2019)Majithia, Arslan, Lubal, Jimenez, Arora,
  Caraballo, and Li}]{majithia-etal-2019-claimportal}
Sarthak Majithia, Fatma Arslan, Sumeet Lubal, Damian Jimenez, Priyank Arora,
  Josue Caraballo, and Chengkai Li. 2019.
\newblock \href {https://doi.org/10.18653/v1/P19-3026} {{C}laim{P}ortal:
  Integrated monitoring, searching, checking, and analytics of factual claims
  on {T}witter}.
\newblock In \emph{Proceedings of the 57th Annual Meeting of the Association
  for Computational Linguistics: System Demonstrations}, pages 153--158,
  Florence, Italy. Association for Computational Linguistics.

\bibitem[{Michael et~al.(2018)Michael, Stanovsky, He, Dagan, and
  Zettlemoyer}]{michael-etal-2018-crowdsourcing}
Julian Michael, Gabriel Stanovsky, Luheng He, Ido Dagan, and Luke Zettlemoyer.
  2018.
\newblock \href {https://doi.org/10.18653/v1/N18-2089} {Crowdsourcing
  question-answer meaning representations}.
\newblock In \emph{Proceedings of the 2018 Conference of the North {A}merican
  Chapter of the Association for Computational Linguistics: Human Language
  Technologies, Volume 2 (Short Papers)}, pages 560--568, New Orleans,
  Louisiana. Association for Computational Linguistics.

\bibitem[{Papadopoulos et~al.(2021)Papadopoulos, Papadakis, and
  Matsatsinis}]{papadopoulos-etal-2021-penelopie}
Dimitris Papadopoulos, Nikolaos Papadakis, and Nikolaos Matsatsinis. 2021.
\newblock \href {https://aclanthology.org/2021.eacl-srw.4} {{PENELOPIE}:
  Enabling open information extraction for the {G}reek language through machine
  translation}.
\newblock In \emph{Proceedings of the 16th Conference of the European Chapter
  of the Association for Computational Linguistics: Student Research Workshop},
  pages 23--29, Online. Association for Computational Linguistics.

\bibitem[{Popat et~al.(2018)Popat, Mukherjee, Yates, and
  Weikum}]{popat-etal-2018-declare}
Kashyap Popat, Subhabrata Mukherjee, Andrew Yates, and Gerhard Weikum. 2018.
\newblock \href {https://doi.org/10.18653/v1/D18-1003} {{D}e{C}lar{E}:
  Debunking fake news and false claims using evidence-aware deep learning}.
\newblock In \emph{Proceedings of the 2018 Conference on Empirical Methods in
  Natural Language Processing}, pages 22--32, Brussels, Belgium. Association
  for Computational Linguistics.

\bibitem[{Radinsky et~al.(2012)Radinsky, Davidovich, and
  Markovitch}]{radinsky2012learning}
Kira Radinsky, Sagie Davidovich, and Shaul Markovitch. 2012.
\newblock Learning to predict from textual data.
\newblock \emph{Journal of Artificial Intelligence Research}, 45:641--684.

\bibitem[{Raffel et~al.(2020)Raffel, Shazeer, Roberts, Lee, Narang, Matena,
  Zhou, Li, and Liu}]{JMLR:v21:20-074}
Colin Raffel, Noam Shazeer, Adam Roberts, Katherine Lee, Sharan Narang, Michael
  Matena, Yanqi Zhou, Wei Li, and Peter~J. Liu. 2020.
\newblock \href {http://jmlr.org/papers/v21/20-074.html} {Exploring the limits
  of transfer learning with a unified text-to-text transformer}.
\newblock \emph{Journal of Machine Learning Research}, 21(140):1--67.

\bibitem[{Reimers and Gurevych(2019)}]{reimers-gurevych-2019-sentence}
Nils Reimers and Iryna Gurevych. 2019.
\newblock \href {https://doi.org/10.18653/v1/D19-1410} {Sentence-{BERT}:
  Sentence embeddings using {S}iamese {BERT}-networks}.
\newblock In \emph{Proceedings of the 2019 Conference on Empirical Methods in
  Natural Language Processing and the 9th International Joint Conference on
  Natural Language Processing (EMNLP-IJCNLP)}, pages 3982--3992, Hong Kong,
  China. Association for Computational Linguistics.

\bibitem[{Reimers and
  Gurevych(2020{\natexlab{a}})}]{reimers-gurevych-2020-making}
Nils Reimers and Iryna Gurevych. 2020{\natexlab{a}}.
\newblock \href {https://doi.org/10.18653/v1/2020.emnlp-main.365} {Making
  monolingual sentence embeddings multilingual using knowledge distillation}.
\newblock In \emph{Proceedings of the 2020 Conference on Empirical Methods in
  Natural Language Processing (EMNLP)}, pages 4512--4525, Online. Association
  for Computational Linguistics.

\bibitem[{Reimers and
  Gurevych(2020{\natexlab{b}})}]{reimers-2020-multilingual-sentence-bert}
Nils Reimers and Iryna Gurevych. 2020{\natexlab{b}}.
\newblock \href {https://arxiv.org/abs/2004.09813} {Making monolingual sentence
  embeddings multilingual using knowledge distillation}.
\newblock In \emph{Proceedings of the 2020 Conference on Empirical Methods in
  Natural Language Processing}. Association for Computational Linguistics.

\bibitem[{Samadi et~al.(2016)Samadi, Talukdar, Veloso, and
  Blum}]{Samadi_Talukdar_Veloso_Blum_2016}
Mehdi Samadi, Partha Talukdar, Manuela Veloso, and Manuel Blum. 2016.
\newblock \href {https://ojs.aaai.org/index.php/AAAI/article/view/9996}
  {Claimeval: Integrated and flexible framework for claim evaluation using
  credibility of sources}.
\newblock \emph{Proceedings of the AAAI Conference on Artificial Intelligence},
  30(1).

\bibitem[{Tang et~al.(2019)Tang, Feng, and Zhao}]{tang-etal-2019-learning}
Jizhi Tang, Yansong Feng, and Dongyan Zhao. 2019.
\newblock \href {https://doi.org/10.18653/v1/D19-1265} {Learning to update
  knowledge graphs by reading news}.
\newblock In \emph{Proceedings of the 2019 Conference on Empirical Methods in
  Natural Language Processing and the 9th International Joint Conference on
  Natural Language Processing (EMNLP-IJCNLP)}, pages 2632--2641, Hong Kong,
  China. Association for Computational Linguistics.

\bibitem[{Thorne et~al.(2018)Thorne, Vlachos, Christodoulopoulos, and
  Mittal}]{thorne-etal-2018-fever}
James Thorne, Andreas Vlachos, Christos Christodoulopoulos, and Arpit Mittal.
  2018.
\newblock \href {https://doi.org/10.18653/v1/N18-1074} {{FEVER}: a large-scale
  dataset for fact extraction and {VER}ification}.
\newblock In \emph{Proceedings of the 2018 Conference of the North {A}merican
  Chapter of the Association for Computational Linguistics: Human Language
  Technologies, Volume 1 (Long Papers)}, pages 809--819, New Orleans,
  Louisiana. Association for Computational Linguistics.

\bibitem[{Vlachos and Riedel(2014)}]{vlachos-riedel-2014-fact}
Andreas Vlachos and Sebastian Riedel. 2014.
\newblock \href {https://doi.org/10.3115/v1/W14-2508} {Fact checking: Task
  definition and dataset construction}.
\newblock In \emph{Proceedings of the {ACL} 2014 Workshop on Language
  Technologies and Computational Social Science}, pages 18--22, Baltimore, MD,
  USA. Association for Computational Linguistics.

\bibitem[{Vossen et~al.(2016)Vossen, Agerri, Aldabe, Cybulska, van Erp,
  Fokkens, Laparra, Minard, Aprosio, Rigau et~al.}]{vossen2016newsreader}
Piek Vossen, Rodrigo Agerri, Itziar Aldabe, Agata Cybulska, Marieke van Erp,
  Antske Fokkens, Egoitz Laparra, Anne-Lyse Minard, Alessio~Palmero Aprosio,
  German Rigau, et~al. 2016.
\newblock Newsreader: Using knowledge resources in a cross-lingual reading
  machine to generate more knowledge from massive streams of news.
\newblock \emph{Knowledge-Based Systems}, 110:60--85.

\bibitem[{Vossen et~al.(2015)Vossen, Caselli, and
  Kontzopoulou}]{vossen-etal-2015-storylines}
Piek Vossen, Tommaso Caselli, and Yiota Kontzopoulou. 2015.
\newblock \href {https://doi.org/10.18653/v1/W15-4507} {Storylines for
  structuring massive streams of news}.
\newblock In \emph{Proceedings of the First Workshop on Computing News
  Storylines}, pages 40--49, Beijing, China. Association for Computational
  Linguistics.

\bibitem[{Webber(2012)}]{10.1145/2384716.2384777}
Jim Webber. 2012.
\newblock \href {https://doi.org/10.1145/2384716.2384777} {A programmatic
  introduction to neo4j}.
\newblock In \emph{Proceedings of the 3rd Annual Conference on Systems,
  Programming, and Applications: Software for Humanity}, SPLASH '12, page
  217–218, New York, NY, USA. Association for Computing Machinery.

\bibitem[{Williams et~al.(2018)Williams, Nangia, and
  Bowman}]{williams-etal-2018-broad}
Adina Williams, Nikita Nangia, and Samuel Bowman. 2018.
\newblock \href {https://doi.org/10.18653/v1/N18-1101} {A broad-coverage
  challenge corpus for sentence understanding through inference}.
\newblock In \emph{Proceedings of the 2018 Conference of the North {A}merican
  Chapter of the Association for Computational Linguistics: Human Language
  Technologies, Volume 1 (Long Papers)}, pages 1112--1122, New Orleans,
  Louisiana. Association for Computational Linguistics.

\bibitem[{Yang et~al.(2019)Yang, Dai, Yang, Carbonell, Salakhutdinov, and
  Le}]{NEURIPS2019_dc6a7e65}
Zhilin Yang, Zihang Dai, Yiming Yang, Jaime Carbonell, Russ~R Salakhutdinov,
  and Quoc~V Le. 2019.
\newblock \href
  {https://proceedings.neurips.cc/paper/2019/file/dc6a7e655d7e5840e66733e9ee67cc69-Paper.pdf}
  {Xlnet: Generalized autoregressive pretraining for language understanding}.
\newblock In \emph{Advances in Neural Information Processing Systems},
  volume~32. Curran Associates, Inc.

\bibitem[{Zeng et~al.(2021)Zeng, Li, Lai, Ji, Bansal, and
  Tong}]{zeng-etal-2021-gene}
Qi~Zeng, Manling Li, Tuan Lai, Heng Ji, Mohit Bansal, and Hanghang Tong. 2021.
\newblock \href {https://aclanthology.org/2021.textgraphs-1.5} {{GENE}: Global
  event network embedding}.
\newblock In \emph{Proceedings of the Fifteenth Workshop on Graph-Based Methods
  for Natural Language Processing (TextGraphs-15)}, pages 42--53, Mexico City,
  Mexico. Association for Computational Linguistics.

\bibitem[{Zhang et~al.(2021)Zhang, Rudra, and Anand}]{10.1145/3459637.3481985}
Zijian Zhang, Koustav Rudra, and Avishek Anand. 2021.
\newblock \href {https://doi.org/10.1145/3459637.3481985} {\emph{FaxPlainAC: A
  Fact-Checking Tool Based on EXPLAINable Models with HumAn Correction in the
  Loop}}, page 4823–4827. Association for Computing Machinery, New York, NY,
  USA.

\bibitem[{Zhou et~al.(2019)Zhou, Han, Yang, Liu, Wang, Li, and
  Sun}]{zhou-etal-2019-gear}
Jie Zhou, Xu~Han, Cheng Yang, Zhiyuan Liu, Lifeng Wang, Changcheng Li, and
  Maosong Sun. 2019.
\newblock \href {https://doi.org/10.18653/v1/P19-1085} {{GEAR}: Graph-based
  evidence aggregating and reasoning for fact verification}.
\newblock In \emph{Proceedings of the 57th Annual Meeting of the Association
  for Computational Linguistics}, pages 892--901, Florence, Italy. Association
  for Computational Linguistics.

\end{thebibliography}
\bibliographystyle{acl_natbib}

\clearpage

    \appendix
    
    \section{Appendix: Original examples (in Greek) of Tables \ref{tab:result1}, \ref{tab:result2} and \ref{tab:result3}.} \label{sec:appendix}

    \begin{table}[hbt!]
        \small
        \begin{tabular}{m{5.3cm}|>{\centering\arraybackslash}m{1.5cm}}
        
            \hline
            \textbf{User Claim \newline \scriptsize (Scenario 1)} & \textbf{NLI score} \\ \hline
            \textgreek{Η \textcolor{Mahogany}{Δανία} και η \textcolor{DarkBlue}{Αυστρία} πιστεύουν ότι η \textcolor{teal}{Ευρωπαϊκή Ένωση} πρέπει να αυξήσει τη βοήθεια προς τους πρόσφυγες.} (\textbf{1a})&
            \cellcolor{Dandelion!20} c: 0.014 \quad \textbf{e: 0.958} \quad n: 0.028 \\ \hline
            \textgreek{Η \textcolor{Mahogany}{Δανία} διαφωνεί με την \textcolor{DarkBlue}{Αυστρία} σχετικά με τη διαχείριση των μεταναστευτικών θεμάτων στην  \textcolor{teal}{Ευρωπαϊκή Ένωση}.} (\textbf{1b}) &
            \cellcolor{Dandelion!20} \textbf{c: 0.951}\quad e: 0.002\quad n: 0.047 \\ \hline
            
            \multicolumn{2}{p{7.2cm}}{\textbf{Candidate Evidence Sequences ($\downarrow$ similarity)}} \\ \hline
            \multicolumn{2}{p{7.2cm}}{\cellcolor{Dandelion!20} \textgreek{Η \textcolor{DarkBlue}{Αυστρία} και η \textcolor{Mahogany}{Δανία} θέλουν να ενισχυθεί επίσης η υποστήριξη της \textcolor{teal}{ΕΕ} προς κράτη που υποδέχονται πρόσφυγες κοντά σε εστίες κρίσεις, ώστε οι πρόσφυγες αυτοί να μην ταξιδεύουν προς την Ευρώπη.} \newline {\scriptsize \textbullet \, \textbf{STS Score: 0.8505}}} \\ \hdashline
            \multicolumn{2}{p{7.2cm}}{\cellcolor{Dandelion!10} \textgreek{Σε έλεγχο από αστυνομικούς των Αστυνομικών Τμημάτων Αερολιμένων ... οι αλλοδαποί επέδειξαν πλαστά ταξιδιωτικά έγγραφα προκειμένου να αναχωρήσουν από τη χώρα για άλλες χώρες της \textcolor{teal}{ΕΕ} όπως η Γαλλία, Γερμανία, Ιταλία, \textcolor{DarkBlue}{Αυστρία}, Ολλανδία, \textcolor{Mahogany}{Δανία}, Ισπανία και Νορβηγία.} \newline {\scriptsize \textbullet \, STS Score: 0.2283}} \\ \hline
            
        \end{tabular}
        \caption{Demonstration of FarFetched on \textit{Scenario 1} in Greek language.}
        \label{tab:result1-greek}
    \end{table}
    
    \begin{table}[hbt!]
        \small
        \begin{tabular}{m{3.4cm}|>{\centering\arraybackslash}m{1.5cm}|>{\centering\arraybackslash}m{1.5cm}}
            \hline
            
            \textbf{User Claim \newline \scriptsize (Scenario 2)} & \textbf{NLI score \scriptsize initial (2a)} & \textbf{NLI score \scriptsize updated (2b)} \\ \hline
            \textgreek{Οι \textcolor{Mahogany}{Ηνωμένες Πολιτείες} σχεδιάζουν να επιβάλλουν κυρώσεις στο \textcolor{DarkBlue}{Ιράν}.} &
            \cellcolor{Dandelion!20} c: 0.170 \quad \textbf{e: 0.571} \quad n: 0.259 &
            \cellcolor{Orchid!20} c: 0.012 \quad \textbf{e: 0.891} \quad n: 0.097 \\ \hline
                
            \multicolumn{3}{p{7.2cm}}{\textbf{Candidate Evidence Sequences ($\downarrow$ similarity)}} \\ \hline
            \multicolumn{3}{p{7.2cm}}{\cellcolor{Dandelion!20}\textgreek{Το \textcolor{DarkBlue}{Ιράν} μπροστά στο δίλημμα αν θα συμμορφωθεί προς τις υποδείξεις της Ουάσινγκτον ή θα οδηγηθεί σε κατάρρευση Οι κυρώσεις που επανήλθαν σε ισχύ σήμερα, θα αναγκάσουν την κυβέρνηση της Ισλαμικής Δημοκρατίας να δεχθεί τις αξιώσεις των \textcolor{Mahogany}{ΗΠΑ} όσον αφορά το ιρανικό πυρηνικό πρόγραμμα και τις ιρανικές δραστηριότητες στην περιοχή της Μέσης Ανατολής διότι, σε διαφορετική περίπτωση, το καθεστώς θα κινδυνεύσει να καταρρεύσει, υποστήριξε ο Ισραέλ Κατς, ο ισραηλινός υπουργός αρμόδιος για τις Υπηρεσίες Πληροφοριών.} \newline {\scriptsize \textbullet \, \textbf{STS Score: 0.6665}}} \\ \hdashline
            \multicolumn{3}{p{7.2cm}}{\cellcolor{Dandelion!15} \textgreek{Γιατί εξαιρέθηκε η Ελλάδα από τις \textcolor{Mahogany}{αμερικανικές} κυρώσεις στο \textcolor{DarkBlue}{Ιράν}. Από τις 5 Νοεμβρίου βρίσκονται σε ισχύ οι νέες κυρώσεις των \textcolor{Mahogany}{ΗΠΑ} για εξαγωγές πετρελαίου από το \textcolor{DarkBlue}{Ιράν}.} \newline {\scriptsize \textbullet \, STS Score: 0.6324}} \\ \hdashline
            \multicolumn{3}{p{7.2cm}}{\cellcolor{Dandelion!10} \textgreek{«Είμαστε πάντα υπέρ της διπλωματίας και των συνομιλιών ... Όμως οι συνομιλίες χρειάζονται εντιμότητα ... Οι \textcolor{Mahogany}{ΗΠΑ} επιβάλλουν εκ νέου κυρώσεις στο \textcolor{DarkBlue}{Ιράν} και αποσύρονται από την πυρηνική συμφωνία (του 2015) και μετά θέλουν να κάνουν συνομιλίες μαζί μας», δήλωσε ο Ροχανί σε ομιλία του που μεταδόθηκε ζωντανά από την τηλεόραση.} \newline {\scriptsize \textbullet \, STS Score: 0.5151}} \\ \hdashline
            \multicolumn{3}{p{7.2cm}}{\cellcolor{Orchid!20} \textgreek{Μετά το ναυάγιο των τελευταίων συνομιλιών μεταξύ \textcolor{Mahogany}{ΗΠΑ} και \textcolor{DarkBlue}{Ιράν} αναμένεται η ανακοίνωση επιπλέον κυρώσεων τις επόμενες ημέρες.} \newline {\scriptsize \textbullet \, \textbf{STS Score: 0.7195}}} \\ \hline
            
        \end{tabular}
        \caption{Demonstration of FarFetched on \textit{Scenario 2} in Greek language.}
        \label{tab:result2-greek}
    \end{table}
    
    \begin{table}[hbt!]
        \small
        \begin{tabular}{m{3.4cm}|>{\centering\arraybackslash}m{1.5cm}|>{\centering\arraybackslash}m{1.5cm}}
            \hline
            
            \textbf{User Claim \newline \scriptsize (Scenario 3)} & \textbf{NLI score \scriptsize initial (3a)} & \textbf{NLI score \scriptsize updated (3b)} \\ \hline
            \textgreek{Η} \textcolor{Mahogany}{Apple} \textgreek{προσπαθεί να ανταγωνιστεί την} \textcolor{DarkBlue}{Netflix} \textgreek{στην παραγωγή τηλεοπτικού περιεχομένου.} &
            \cellcolor{Dandelion!20} c: 0.004 \quad \textbf{e: 0.967} \quad n: 0.029 &
            \cellcolor{Orchid!20} \textbf{c: 0.982} \quad e: 0.008 \quad n: 0.010 \\ \hline
            
            \multicolumn{3}{p{7.2cm}}{\textbf{Candidate Evidence Sequences ($\downarrow$ similarity)}} \\ \hline
            \multicolumn{3}{p{7.2cm}}{\cellcolor{Dandelion!20} \textgreek{Η} \textcolor{Mahogany}{Apple} \textgreek{αναμένεται να δαπανήσει φέτος περίπου 2 δισεκατομμύρια δολάρια με σκοπό τη δημιουργία πρωτότυπου περιεχομένου που ελπίζει ότι θα ανταγωνιστεί τις ήδη εδραιωμένες στο τηλεοπτικό κοινό υπηρεσίες των} \textcolor{DarkBlue}{Netflix}, Hulu \textgreek{και} Amazon\textgreek{.} \newline {\scriptsize \textbullet \, \textbf{STS Score: 0.7107}}} \\ \hdashline
            \multicolumn{3}{p{7.2cm}}{ \cellcolor{Orchid!20} \textgreek{«Δεν προσπαθούμε να ανταγωνιστούμε το} \textcolor{DarkBlue}{Netflix} \textgreek{στην τηλεόραση», δήλωσε εκπρόσωπος της} \textcolor{Mahogany}{Apple} \textgreek{σε συνέντευξή του.} \newline {\scriptsize \textbullet \, \textbf{STS Score: 0.7134}}} \\ \hline
            
        \end{tabular}
        \caption{Demonstration of FarFetched on \textit{Scenario 3} in Greek language.}
        \label{tab:tab:result3-greek}
    \end{table}

    \end{document}